\documentclass[letter, 10 pt, conference]{ieeeconf}  

\IEEEoverridecommandlockouts

\usepackage{mathptmx}       
\usepackage{helvet}         
\usepackage{courier}        
\usepackage{type1cm}        
\usepackage{algorithm}
\usepackage[noend]{algorithmic}
\usepackage{siunitx}
\usepackage{colortbl}
\usepackage{array}
\usepackage{booktabs}
\usepackage{amsmath}
\usepackage{amsfonts}
\usepackage{blindtext}
\usepackage{bm}
\usepackage{subfigure}                            
\usepackage{makeidx}         
\usepackage{graphicx}        
\usepackage{multicol}        
\usepackage[bottom]{footmisc}
\usepackage{comment}
\usepackage{balance}
\usepackage[mathcal]{eucal}
\DeclareSymbolFont{usualmathcal}{OMS}{cmsy}{m}{n}
\DeclareSymbolFontAlphabet{\mathcal}{usualmathcal}

\usepackage{dblfloatfix}
\usepackage{lmodern}
\usepackage{cite}
\let\labelindent\relax
\usepackage{enumitem}
\usepackage[hyphens]{url}

\title{\LARGE \bf Perception-and-Energy-aware Motion Planning for UAV \\ using Learning-based Model under Heteroscedastic Uncertainty}
\author{Reiya Takemura$^{1}$ and Genya Ishigami$^{1}$
\thanks{The authors are with the Graduate School of Integrated Design Engineering, Keio University, Japan (e-mail: {\tt\small rereon@keio.jp}; {\tt\small ishigami@mech.keio.ac.jp}).
}}

\begin{document}
\newcommand{\bhline}[1]{\noalign{\hrule height #1}}
\newcommand{\bvline}[1]{\vrule width #1}

\onecolumn{
© 2023 IEEE. Personal use of this material is permitted. Permission from IEEE must be obtained for all other uses, in any
current or future media, including reprinting/republishing this material for advertising or promotional purposes, creating
new collective works, for resale or redistribution to servers or lists, or reuse of any copyrighted component of this work in
other works.
}

\newpage
\twocolumn
\maketitle
\thispagestyle{empty}
\pagestyle{empty}

\begin{abstract}
  Global navigation satellite systems (GNSS) denied environments/conditions require unmanned aerial vehicles (UAVs) to energy-efficiently and reliably fly.
  To this end, this study presents perception-and-energy-aware motion planning for UAVs in GNSS-denied environments.
  The proposed planner solves the trajectory planning problem by optimizing a cost function consisting of two indices: the total energy consumption of a UAV and the perception quality of light detection and ranging (LiDAR) sensor mounted on the UAV.
  Before online navigation, a high-fidelity simulator acquires a flight dataset to learn energy consumption for the UAV and heteroscedastic uncertainty associated with LiDAR measurements, both as functions of the horizontal velocity of the UAV.
  The learned models enable the online planner to estimate energy consumption and perception quality, reducing UAV battery usage and localization errors.
  Simulation experiments in a photorealistic environment confirm that the proposed planner can address the trade-off between energy efficiency and perception quality under heteroscedastic uncertainty.
  The open-source code is released at \url{https://gitlab.com/ReI08/perception-energy-planner}.
\end{abstract}
\section{INTRODUCTION}
Autonomous unmanned aerial vehicles (UAVs) have been widely used for various missions such as construction/infrastructure inspection and search and rescue in disaster environments.
In these missions, UAVs are often limited in their flight distance/duration because of the shortage of onboard batteries or positioning errors owing to onboard sensors and loss of signal from global navigation satellite systems (GNSSs).

\begin{figure}[!t]
\centering
\includegraphics[width=0.8\linewidth]{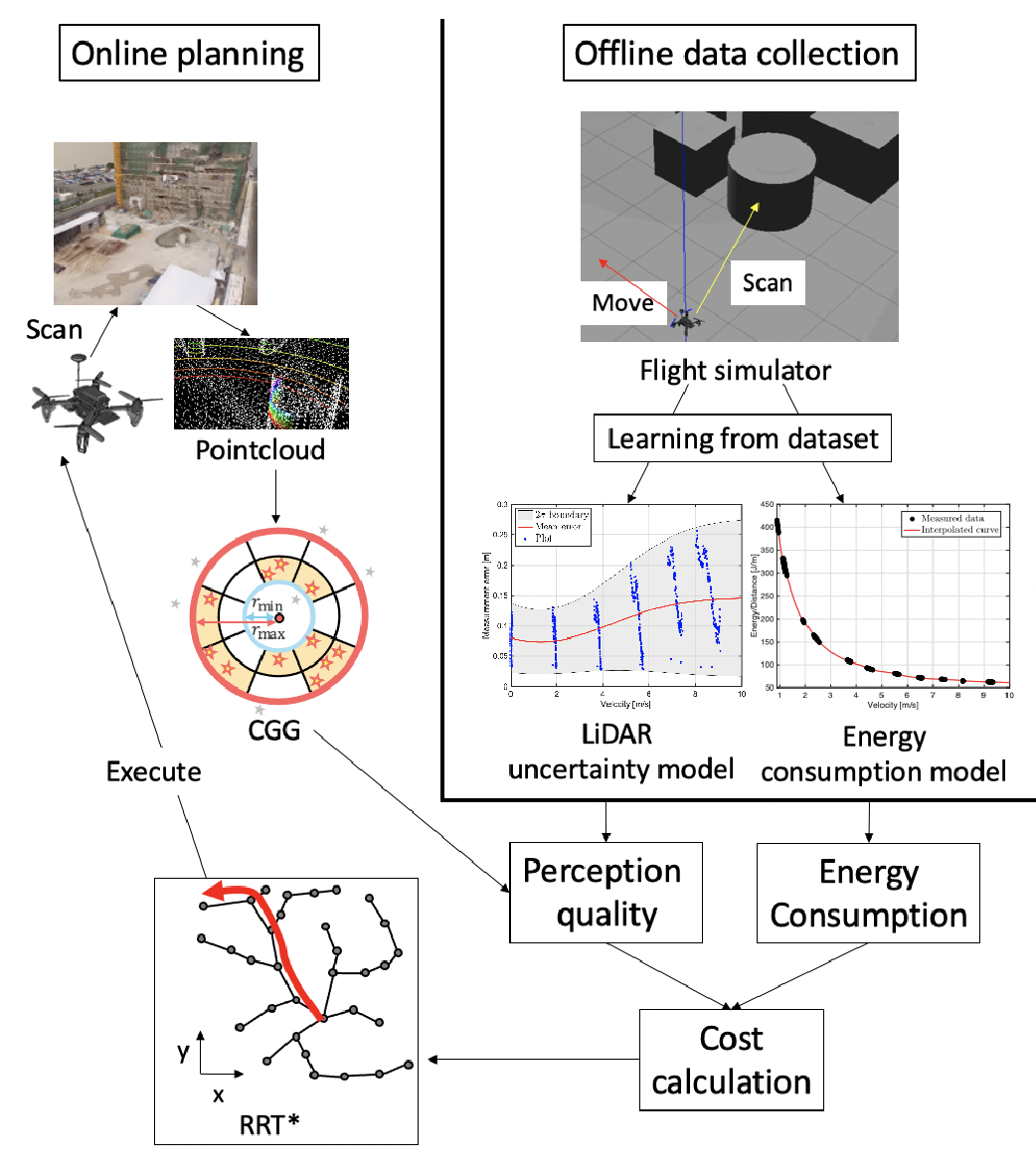}
\caption{Framework for the Perception-and-Energy-aware motion planning. High-fidelity simulator collects flight data to learn light detection and ranging (LiDAR) uncertainty and energy consumption models (right). Using the models, the online planner calculates energy consumption and perception quality on a circular grid graph and generates a perception-and-energy-aware trajectory (left).}
\label{fig:overview}
\end{figure}

To address the battery problem, a UAV should be lightened to the extent possible while multiple battery packs should be mounted on the UAV for long-range flight, increasing the weight of the UAV.
The trade-off between the UAV's weight and its batteries severely limits the flight time and range.
To address this issue, recent studies have proposed energy-efficient motion planning using power consumption models of UAVs.
The common approaches for modeling the power consumption of UAVs include theoretical and data-driven approaches.
The former describes the physical phenomenon of UAVs based on aerodynamics and rotor/battery models \cite{Abdilla2015, Bezzo2016, Tagliabue2019}.
These works calculate the thrust of UAVs or use the power-thrust curve of rotors to estimate power consumption.
The theoretical model cannot easily optimize the coefficients and parameters (e.g., the drag coefficient and rotor efficiency) because they vary with different types of drones and flight conditions.
In contrast, the latter can easily be implemented without extensive calculations of aerodynamics, while a considerable amount of experimental flight data for power consumption is required \cite{DiFranco2016, Abeywickrama2018a, Baek2019, Hong2020}.
These works can predict energy consumption with high accuracy if variables such as the velocity and acceleration of UAVs are effectively exploited for model inputs.

GNSS-denied environments/conditions often observed in stormy mountainous areas, forests, and even urban canyons require UAVs to reliably fly autonomously.
Simultaneous localization and mapping (SLAM) is capable of self-pose estimation as it utilizes light detection and ranging (LiDAR) that directly measures three-dimensional distances from the sensor to the objects with a wide field of view.
Several studies have developed SLAM methods to reduce the state estimation error of UAVs in GNSS-denied environments \cite{Zhang2014, Ye2019, Tixiao2020}.
However, these studies have several limitations. For instance, pose estimation performs poorly in large measurement uncertainty or non-landmark feature points.
This is because LiDAR-based SLAM suffers from sensing noise and geometric degenerations, such as long straight tunnels and planar environments.
A few studies have addressed the degeneration problem for the LiDAR-based SLAM \cite{Zhang2016, Zhen2019, Jiao2021, Zhou2021}.
To resolve the optimization problem for pose estimation, Zhang et al. used only well-conditioned constraints, and Jiao et al. selected valuable/informative features.
Zhou et al. exploited the Fisher information matrix (FIM) associated with LiDAR measurement uncertainty, thereby detecting the degeneration degree of sensor measurements.
Zhang et al. proposed a perception-aware path planning method to address such degeneration problem \cite{Zhang2020}.
They introduced the idea of the Fisher information field (FIF) to efficiently calculate the FIM from a set of a sparse point cloud.
Although a planner with FIF increases the localization success rate and accuracy, it is restricted to known environments because the accurate pre-computation of the FIF is needed.
Our previous study also proposed a perception-aware path planner, which assesses the perception quality for a LiDAR-based SLAM algorithm online \cite{Takemura2022a}.

Thus far, several studies on energy-efficient or perception-aware motion planning have been reported. However, none have focused on both simultaneously.
This issue may sometimes be involved in the trade-off between energy efficiency and perception quality.
Intuitively, as the velocity of a UAV increases, the perception quality for SLAM likely decreases.
Conversely, at a lower velocity, the total energy consumption increases as reported in the literature \cite{DiFranco2016, Hong2020}.
Further, the FIM calculation for the perception quality \cite{Zhang2020, Zhou2021} is demonstrated to be valid only when LiDAR measurement uncertainty is constant.
However, the measurement uncertainty is dependent on several factors, such as UAV behaviour, leading to heteroscedastic uncertainty.
Therefore, this study proposes perception-and-energy-aware motion planning for UAVs under heteroscedastic uncertainty assumption, enabling them to fly while maintaining energy efficiency and pose estimation accuracy.
The proposed planning framework consists of offline and online components, as shown in Fig. \ref{fig:overview}.
In the offline process, a high-fidelity simulator collects a flight dataset to learn heteroscedastic uncertainty for LiDAR measurements and the energy consumption of a UAV according to its velocity.
Using learned models, the online planner solves the trajectory planning problem to minimize the cost function, which comprised the perception quality and total energy consumption.
Our research highlights are as follows:
\begin{itemize}
\item Learning heteroscedastic uncertainty in LiDAR measurements correlated with the velocity of a UAV;
\item Designing a trajectory planner, which optimize the trade-off between the energy consumption and perception quality using the LiDAR uncertainty model;
\item Verifying the proposed planner can solve the trade-off under heteroscedastic uncertainty; and
\item Validating the proposed planner using photorealistic simulations.
\end{itemize}

The remainder of this paper is organized as follows: Section I\hspace{-.1em}I defines the problem statement. Section I\hspace{-.1em}I\hspace{-.1em}I presents the proposed motion planning with the energy consumption and perception quality for LiDAR uncertainty. Section I\hspace{-.1em}V discusses the results of our simulation experiment. Finally, Section V presents the conclusion and scope for future studies.

\begin{figure}[t]
\centering
\includegraphics[width=0.45\linewidth]{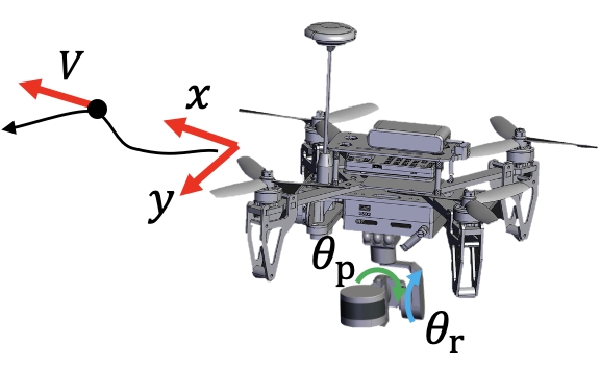}
\caption{Schematic view of the UAV with the gimbal-based 3D LiDAR system.}
\label{fig:vehicle}
\end{figure}

\newcommand{\argmax}{\mathop{\rm arg~max}\limits}
\newcommand{\argmin}{\mathop{\rm arg~min}\limits}
\section{PROBLEM STATEMENT}
\subsection{System for UAV with gimbal-based 3D LiDAR}
As shown in Fig. \ref{fig:vehicle}, we assume that a UAV is equipped with a 3D LiDAR sensor on its bottom, enabling 3D scanning with a 360$^\circ$ field of view.
The state variable of the UAV is expressed as follows:
\begin{equation}
\label{state}
\bm{x} = [x \ y \ V \ \theta_{\rm{r}} \ \theta_{\rm{p}}]^\mathrm{T},
\end{equation}
where $x$ and $y$ are the coordinates of the center of gravity of the UAV.
$V$ is the horizontal velocity of the UAV and operates along the tangential direction of the $x$-$y$ curve.
We assume that the UAV moves horizontally because vertical takeoff and landing UAVs are capable of efficient exploratory flight.
$\theta_{\rm{r}}$ and $\theta_{\rm{p}}$ are the gimbal angles, enabling the LiDAR sensor to scan the terrain surface horizontally even if the UAV is tilted;
otherwise, the vertical field of view of LiDAR would be relatively narrow.
The gimbal automatically controls $\theta_{\rm{r}}$ and $\theta_{\rm{p}}$ to be inverse angles to the posture of the UAV obtained from an inertia measurement unit mounted on the UAV. Therefore, these variables are ignored in the motion planning process.

\subsection{System dynamics and observation model}
Generally, using the system dynamics $f(\cdot)$ of a UAV with an observation model ${}^kh(\cdot)$ for the $k$-th feature point of the SLAM algorithm, the state and sensor measurement of the UAV are described in a discrete-time manner as follows:
\begin{eqnarray}
\label{eq:system_dynamics1}
\bm{x}_{n} = f(\bm{x}_{n-1}, \bm{u}_{n-1}), \ \ \ \bm{x}_{n} \in \bm{X}, \ \bm{u}_n \in \bm{U},
\end{eqnarray}
\begin{eqnarray}
\label{eq:system_dynamics2}
{}^kz_{n} = {}^kh(\bm{x}_{n}) + {}^k\xi_{n},
\end{eqnarray}
where the state variable $\bm{x}_{n}$ and control variable $\bm{u}_n$ describe the time evolution of the system dynamics from time $n-1$ to $n$.
$\bm{X}$ and $\bm{U}$ are the state and control input spaces of the UAV, respectively.
${}^kz_{n}$ is the measurement of the $k$-th feature point.
In LiDAR-based observation, the number of feature points is quite numerous; therefore, the LiDAR odometry and mapping (LAOM) algorithm is exploited as the SLAM to estimate the state using ${}^kz_{n}$ \cite{Zhang2014}.
${}^k\xi_{n}$ is the observation noise obtained using a Gaussian distribution with the absolute error $\mu_{\rm{sens}}$ and repeatability $\sigma_{\rm{sens}}$, which may cause the degeneration of the LOAM algorithm.
According to \cite{Tanaka2016}, the absolute error is based on the residual between the mean of the measured values and true value, and the repeatability expresses the standard deviation for sensor measurements.
In this study, the measurement probability is assumed to be a Gaussian distribution; hence the sensor model is described using the following equation:
\begin{equation}
\label{eq:sensor_model}
p({}^kz_{n}|\bm{x}_n) = \frac{1}{\sqrt{2\pi\sigma^2_{\rm{sens}}}}\exp{-\frac{\left({}^kz_{n}-({}^kh(\bm{x}_{n}) + \mu_{\rm{sens}})\right)^2}{2\sigma^2_{\rm{sens}}}}.
\end{equation}

We assume that the observation uncertainty $\mu_{\rm{sens}}$ and $\sigma_{\rm{sens}}$ are decoupled into two, derived from the motion of the UAV and the distance between the UAV and scanned object:
\begin{eqnarray}
\label{eq:sigma}
\mu_{\rm{sens}} = \mu_{\rm{D}} + \mu_{\rm{M}}, \ \ \ \ \sigma^2_{\rm{sens}} = \sigma^2_{\rm{D}} + \sigma^2_{\rm{M}},
\end{eqnarray}
where $\mu_{\rm{D}}$ and $\sigma_{\rm{D}}$ describe distance-based uncertainty, and
$\mu_{\rm{M}}$ and $\sigma_{\rm{M}}$ denote motion-based uncertainty.
This assumption is derived from the literature \cite{Tanaka2016, Schlager2022a}, where LiDAR uncertainty is affected by long-range measurements and vibrations from the flight motions of UAVs.
For distance-based uncertainty, $\sigma^2_{\rm{D}}$ can be quantified using the experimental results in \cite{Kidd2017}.
Notably, this study ignores $\mu_{\rm{D}}$ as it is easily offset by the LiDAR calibration.
For motion-based uncertainty, we assume heteroscedastic uncertainty, meaning that $\mu_{\rm{M}}$ and $\sigma_{\rm{M}}$ are dependent on the horizontal velocity of the UAV.
This is because vibrations for the LiDAR measurement are caused by the gimbal control angles $\theta_{\rm{r}}$ and $\theta_{\rm{p}}$, which increase when the UAV flies rapidly and tilts significantly.
By preparing a dataset of UAV velocity versus observation error from LiDAR measurement experiments, observation uncertainty is modeled using a learning-based regression function.
$\mu_{\rm{M}}$ and $\sigma_{\rm{M}}$ increase in proportion to the velocity of the UAV because the gimbal tilts and vibrates extensively during its fast flights, which enlarges LiDAR measurement uncertainty.

\subsection{Objective for motion planning}
This study ensures that a UAV reaches a given goal area in GNSS-denied and unknown environments while reducing pose error and energy consumption.
We address this challenge as motion planning, where the planner finds a feasible trajectory to a goal region.
This motion planning determines sequential states with the perception quality and energy efficiency of the UAV as follows:
\begin{eqnarray}
\label{eq:optimize_cost}
\mathcal{T^*} = \argmin_{\mathcal{T}=[\bm{x}_{\rm{s}}, ..., \bm{x}_{\rm{g}}]} \{\alpha_{\rm{P}} C_{\rm{P}}(\mathcal{T})+ \alpha_{\rm{E}} C_{\rm{E}}(\mathcal{T})\},
\end{eqnarray}
where $C_{\rm{P}}$ and $C_{\rm{E}}$ are the cost related to perception quality and energy consumption, respectively.
The parameters $\alpha_{\rm{P}}$ and $\alpha_{\rm{E}}$ are the weighting factor to prioritize each term.
$\mathcal{T}$ is the nominal trajectory, which is given by the sequence of the UAV's state from the initial state $\bm{x}_{\rm{s}}$ to the goal state $\bm{x}_{\rm{g}}$ in $\bm{X}_{\rm{goal}}$.

Summarily, our planning problem can be defined as follows:
Given $\bm{x}_{\rm{s}}$ and $X_{\rm{goal}}$, \ solve Eq. (\ref{eq:optimize_cost}) \ \ s.t. Eqs. (\ref{eq:system_dynamics1})-(\ref{eq:sigma}).
\vspace{-1mm}
\section{MOTION PLANNING ALGORITHM}
\subsection{Overview}
The proposed motion planning algorithm searches for a perception-aware trajectory with lower energy consumption.
To this end, we exploit the framework of a rapidly-exploring random tree star (RRT$^*$) \cite{Karaman2011} with a spline smoothing approach.
An RRT$^*$ algorithm can generate a cost-optimal path \cite{Karaman2011}. It has widely been used for the motion planning of UAVs.
Despite its advantage, RRT$^*$-based planners in high-dimensional problems are still time-consuming because numerical simulation with a dynamic model requires a high computational burden to propagate the tree structure.
Therefore, a spline curve is used for path smoothing while the RRT$^*$ algorithm extends its tree on the $x$-$y$ plane.
Cubic spline interpolation exploits the extended tree as control points, as reported in the literature \cite{Lee2014}.

Algorithm \ref{alg1} shows the procedure of our algorithm.
Before starting the motion planning, $G_{\rm{tree}}$ is defined as a tree using the initial state (Alg. \ref{alg1}, Line 1).
The Sample function randomly samples a node including $x$ and $y$ coordinates (Alg. \ref{alg1}, Line 3).
The GetNearTipStates function searches for states near the extended node among tip states of each trajectory segment in $G_{\rm{tree}}$ (Alg. \ref{alg1}, Line 6).
In the loop (Alg. \ref{alg1}, Line 7), the tree attempts to extend from the near state to the new node and spline-based smoothing outputs the $x$-$y$ curve described by $S_x$ and $S_y$ (Alg. \ref{alg1}, Line 8).
Subsequently, the velocity is optimized (Alg. \ref{alg1}, Line 9), where multiple velocity commands are tried in the range of $V_{\rm{min}}$ to $V_{\rm{max}}$, and the cost consisting of energy consumption and perception quality is compared.
Then, the trajectory segment $(\bm{x}_{i}, \bm{x}_{i+1}, ... \bm{x}_{\rm{tip}})$, which can minimize the cost from $\bm{x}_{\rm{s}}$ to $\bm{x}_{\rm{tip}}$ through $\bm{x}_{\rm{near}}$, is selected (Alg. \ref{alg1}, Lines 10-13) and added to the tree (Alg. \ref{alg1}, Line 14).
As in a basic RRT$^*$ algorithm, a rewiring process is performed (Alg. \ref{alg1}, Lines 15-17).
The above-mentioned procedures are repeated until the tree reaches the goal and the number of sampled nodes reaches its threshold (Alg. \ref{alg1}, Line 18).
The following subsection describes the calculation of the costs $C_{\rm{P}}$ and $C_{\rm{E}}$ related to the perception quality and energy consumption, respectively.
\vspace{-1mm}
\renewcommand{\algorithmicrequire}{\textbf{Input:}}
\renewcommand{\algorithmicensure}{\textbf{Output:}}
\begin{center}
\begin{frame}{}
\scalebox{0.7}{
\begin{minipage}{1.1\linewidth}
\begin{algorithm}[H]
\caption{Perception-and-Energy-aware Motion Planning}
\label{alg1}
\begin{algorithmic}[1]
  \REQUIRE Initial state: $\bm{x}_{\rm{s}}$\\
  \REQUIRE Goal region: $X_{\rm{g}}$\\
  \ENSURE Trajectory: $\cal{T}$\\
\STATE $G_{\rm{tree}}\leftarrow$Initialize($\bm{x}_{\rm{s}}$)
\REPEAT
\STATE ($x_{\rm{rand}}, y_{\rm{rand}})\leftarrow$Sample($X_{\rm{goal}}$)
\STATE ($x_{\rm{nearest}}, y_{\rm{nearest}})\leftarrow$GetNearest($G_{\rm{tree}}, x_{\rm{rand}}, y_{\rm{rand}}$)
\STATE $(x_{\rm{new}}, y_{\rm{new}})\leftarrow$Extend($x_{\rm{nearest}}, y_{\rm{nearest}}, x_{\rm{rand}}, y_{\rm{rand}}$)
\STATE $X_{\rm{near}}\leftarrow$GetNearTipStates($G_{\rm{tree}}, x_{\rm{new}}, y_{\rm{new}}$)
\FOR{$\bm{x}_{\rm{near}}$ in $X_{\rm{near}}$}
\STATE $(S_{x}, S_{y})\leftarrow$Smooth($\bm{x}_{\rm{near}}, x_{\rm{new}}, y_{\rm{new}}$)
\STATE (($\bm{x}_{i}, \bm{x}_{i+1}, ... \bm{x}_{\rm{tip}}$)$, C$)$\leftarrow$OptVel$(S_{x}, S_{y}, V_{\bm{x}_{\rm{near}}}$)
 \IF {$C < C_{\rm{new}}$}
    \STATE $\bm{x}_{i_{\rm{parent}}}\leftarrow\bm{x}_{\rm{near}}$
		\STATE $X_{\rm{new}}\leftarrow$ ($\bm{x}_{i}, \bm{x}_{i+1}, ... \bm{x}_{\rm{tip}}$)
    \STATE $C_{\rm{new}}\leftarrow C$
	\ENDIF
\ENDFOR
\STATE $G_{\rm{tree}}\leftarrow$AddTrajSeg($X_{\rm{new}}$)
\STATE $X_{\rm{near}}\leftarrow$GetNearTipStates($G_{\rm{tree}}, x_{\bm{x}_{\rm{tip}}}, y_{\bm{x}_{\rm{tip}}}$)
\FOR{$\bm{x}_{\rm{near}}$ in $X_{\rm{near}}$}
\STATE $G_{\rm{tree}}\leftarrow$Rewire($\bm{x}_{\rm{tip}}, \bm{x}_{\rm{near}}$)
\ENDFOR
 \UNTIL{Tree $G_{\rm{tree}}$ reaches goal region $X_{\rm{goal}}$ and the number of sampling reaches threshold}
 \STATE $\cal{T}\leftarrow$GetTrajectory($G_{\rm{tree}}$)
  \STATE Return $\cal{T}$
\end{algorithmic}
\end{algorithm}
\end{minipage}
}
\end{frame}
\end{center}

\subsection{Cost for Perception Quality}
Recently, as in the literature \cite{Zhang2020, Zhou2021}, the FIM has been combined with motion planners, leading to the selection of rich information from sensor observations in parameter estimation problems such as SLAM.
This study also exploits the FIM to calculate the cost related to the perception quality for the trajectory segment $(\bm{x}_{i}, \bm{x}_{i+1}, ... \bm{x}_{\rm{tip}})$:
\begin{equation}
  \label{eq:C_P}
  C_{\rm{P}} = \int_{t_i:\bm{x}_{i}}^{t_{\rm{tip}}:\bm{x}_{\rm{tip}}}\frac{1}{s(\rm{I}_{\bm{x}})}dt, \ \ \rm{I}_{\bm{x}} = \sum_{\it{j}}^{\it{j}\in G_{\bm{x}}}\rm{I}_{\it{g_j}},
\end{equation}
where $\rm{I}_{\bm{x}}$ is each FIM for each state $\bm{x}$.
$s(\cdot)$ is a metric function such as the determinant and the smallest eigenvalue, allowing the FIM to be a scalar.
$G_{\bm{x}}$ denotes a circular grid graph (CGG) at the state $\bm{x}$, and $g_j$ is the $j$-th true grid in the CGG.
All the grids in the CGG are constructed by the feature points extracted in the LOAM algorithm.
For each grid, the FIM is calculated based on $g_j$ and $\bm{x}$.
\subsubsection{Fisher information matrix on CGG}
\begin{figure}[t]
\centering
\includegraphics[width=0.65\linewidth]{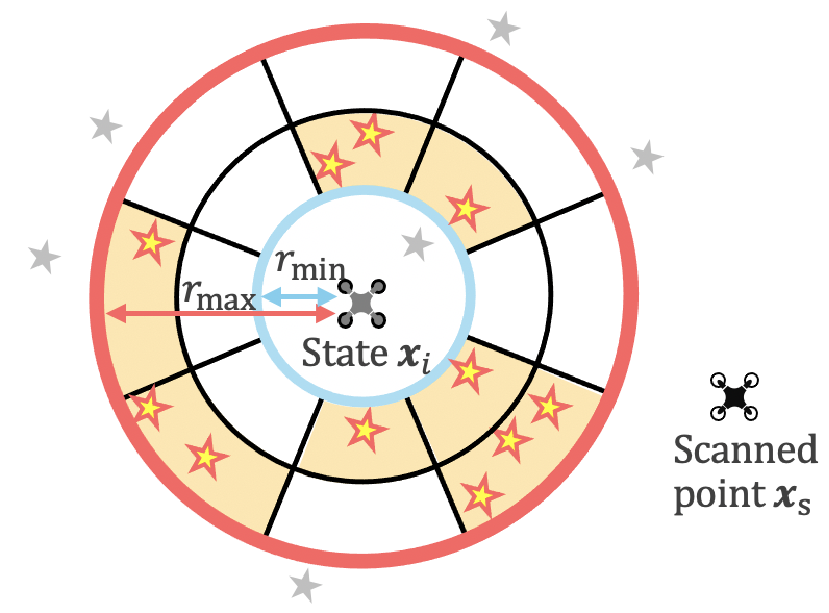}
\caption{CGG for the FIM calculation at the state $\bm{x}_i$. The LiDAR-based SLAM algorithm extracts feature points (star shaped markers) from the point cloud that the UAV acquires at the scanned point. A grid in the graph becomes true (orange) if it contains at least one feature point.}
\label{fig:circ_grid}
\end{figure}

$\rm{I}_{\bm{x}}$ needs to be calculated for numerous feature points that would be visible from the UAV if the CGG were not used.
This causes a time-consuming motion-planning problem.
Therefore, we introduce the graph architecture, as shown in Fig. \ref{fig:circ_grid}, which was first developed in our previous study \cite{Takemura2022a}.
The graph is used for the downsampling of the numerous feature points, decreasing the calculation of the FIM.
As shown in Fig. \ref{fig:circ_grid}, the graph contains grids in radial and angular directions. Each grid is true when it contains at least one feature point of the LOAM algorithm.
The CGG is defined using two radii $r_{\rm{min}}$ and $r_{\rm{max}}$.
$r_{\rm{min}}$ is the minimum radius for obstacle avoidance because feature-rich objects near the UAV are at risk of collision.
$r_{\rm{max}}$ is based on the reliable measurement range of LiDAR because feature points far from the UAV lack its accuracy.
Notably, the FIM is calculated only for each true grid in the CGG.
Compared with the literature \cite{Zhang2020}, this method is useful even for an unknown environment because the motion planner does not need to prepare the map.

Subsequently, let us introduce the FIM for the pseudo-measurement of the grid $g$ in the CGG.
Generally, the FIM is defined as follows:
\begin{equation}
  \label{eq:fi_ori}
\rm{I}_{\it{g}} = E\left[(\frac{\partial}{\partial \bm{x}}\log{\it{p}(g|\bm{x})})^\top(\frac{\partial}{\partial \bm{x}}\log{\it{p}(g|\bm{x})})\right],
\end{equation}
where $p$ is the measurement probability, which is described by Eq. (\ref{eq:sensor_model}).
As in the reference \cite{Steven1993}, Eq. (\ref{eq:fi_ori}) can be transformed as follows:
\begin{eqnarray}
  \label{eq:fi}
\rm{I}_{\it{g}} = \frac{\rm{1}}{\sigma^{\rm{2}}_{\rm{sens}}}({}^1\rm{J}_{\bm{x}})^\top {}^1\rm{J}_{\bm{x}} + \frac{1}{2\sigma^4_{\rm{sens}}}({}^2\rm{J}_{\bm{x}})^\top {}^2\rm{J}_{\bm{x}}, \\
{}^1\rm{J}_{\bm{x}} = \frac{\partial ({}^{\it g}\it{h}(\bm{x})+\mu_{\rm{sens}})}{\partial \bm{x}}, \ \ \ {}^2J_{\bm{x}} = \frac{\partial \sigma_{\rm{sens}}}{\partial \bm{x}} \nonumber.
\end{eqnarray}
Once $\mu_{\rm{sens}}$ and $\sigma_{\rm{sens}}$ are experimentally learned in the offline process, the FIM can be calculated even in online motion planning.

Notably, the FIM is a vital concept in parameter estimation theory and statistics.
Recently, the FIM is used to determine the degeneration degree of the LiDAR measurements for the LOAM algorithm \cite{Zhang2020, Zhou2021}.
However, they demonstrated only when LiDAR measurement uncertainty is constant.
In this study, the FIM with the heteroscedastic uncertainty is used to express the perception quality and this is verified.
\subsubsection{LiDAR measurement experiment}
\begin{figure}[t]
\centering
\begin{minipage}[t]{0.38\linewidth}
\centering
\includegraphics[width=1.0\linewidth]{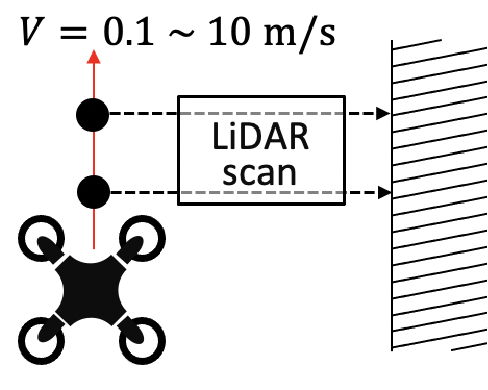} \\
{\centering(a) }
\end{minipage}
\begin{minipage}[t]{0.6\linewidth}
\centering
\includegraphics[width=1.0\linewidth]{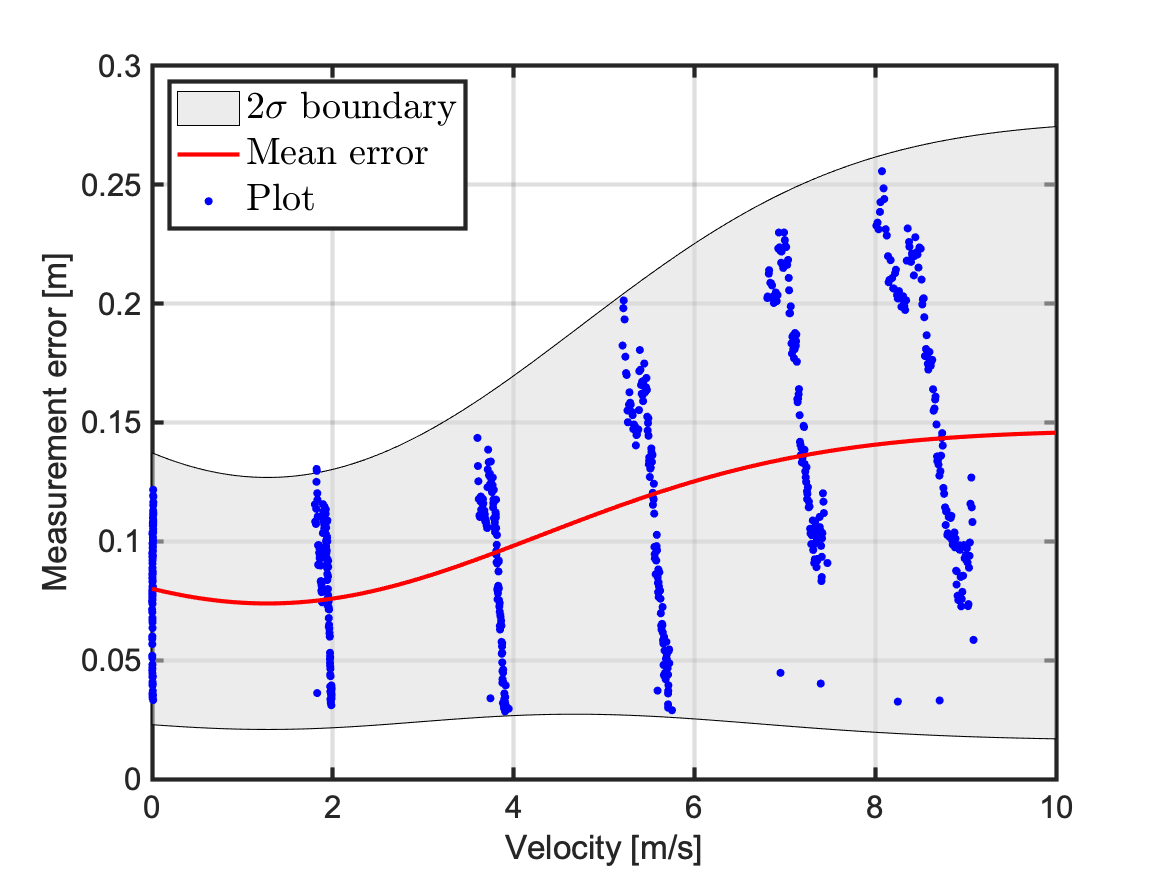} \\
{\centering(b)}
\end{minipage}
\caption{Modeling motion-based uncertainty for gimbal-based LiDAR system. (a) Illustration of LiDAR measurement experiment. (b) Measurement results and learned model by HGPR.}
\label{fig:lidar_exp_unc_model}
\end{figure}

For modeling the motion-based uncertainty of LiDAR measurements, we prepared high-fidelity simulation environments in Gazebo supported with the PX4 software in the loop simulation. 
The UAV model is an extension of PX4 Vision, replacing its camera with a gimbal-based 3D LiDAR (VLP-16) simulation model \cite{velodyne}, as shown in Fig. \ref{fig:vehicle}.
In the simulation, the UAV moves straight ahead with a constant velocity for several seconds and the LiDAR sensor scans along the wall, as shown in Fig. \ref{fig:lidar_exp_unc_model} (a).
The origin of the scanned point cloud is converted to the center of gravity of the UAV, using the gimbal angle $\theta_{\rm{r}}$ and $\theta_{\rm{p}}$.
The Gazebo simulator can log the actual distance from the scanned point to the UAV, enabling the calculation of the LiDAR measurement residual between the true and measured values during the flight.
The velocity setting of the UAV is in the range of [0.1, 2.0, 4.0, 6.0, 8.0, 10.0] m/s.
Notably, the field experiment using an actual UAV could also be used to model the LiDAR measurement uncertainty.
However, we could not determine whether the experimental results can be derived from either distance-based or motion-based uncertainty. Conversely, the Gazebo simulator can set the distance-based uncertainty as zero during measurements.
Therefore, this study exploited the simulation environment.

Figure \ref{fig:lidar_exp_unc_model} (b) shows the simulation result for LiDAR measurement uncertainty.
As we expected, the measurement error and its variance increase when the velocity of the UAV increases.
Subsequently, the heteroscedastic Gaussian processes regression (HGPR) \cite{Almosallam2017} is exploited to learn the dataset.
In this figure, the mean error shows the absolute error $\mu_{\rm{M}}$ and $\sigma$ boundary describes the repeatability $\sigma_{\rm{M}}$.
Notably, the learned model is used to calculate the FIM in Eq. (\ref{eq:fi}).

\vspace{-2mm}
\subsection{Cost for Energy Consumption}
\begin{figure}[t]
\centering
\hspace{-3mm}
\begin{minipage}[t]{0.49\linewidth}
\centering
\includegraphics[width=1.1\linewidth]{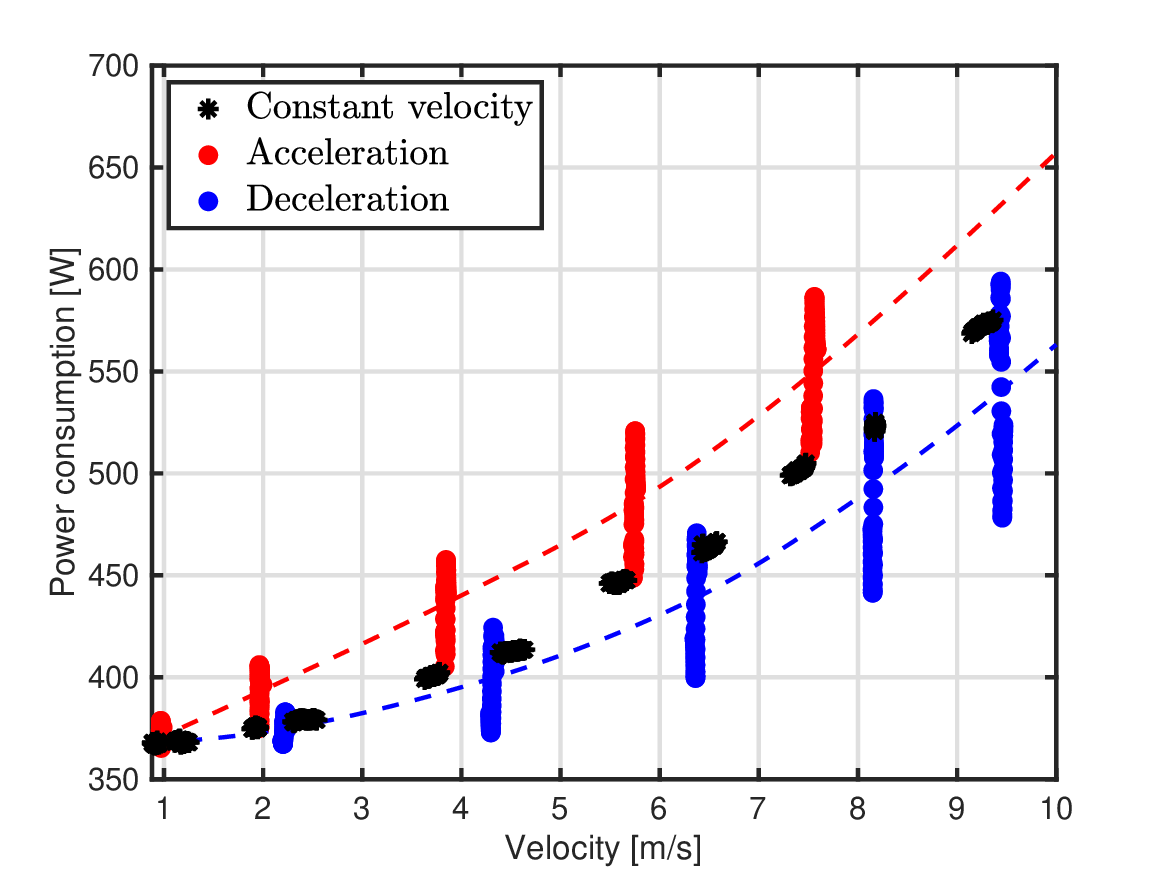}\\
  {\centering (a)}
\end{minipage}
\begin{minipage}[t]{0.49\linewidth}
\centering
\includegraphics[width=1.1\linewidth]{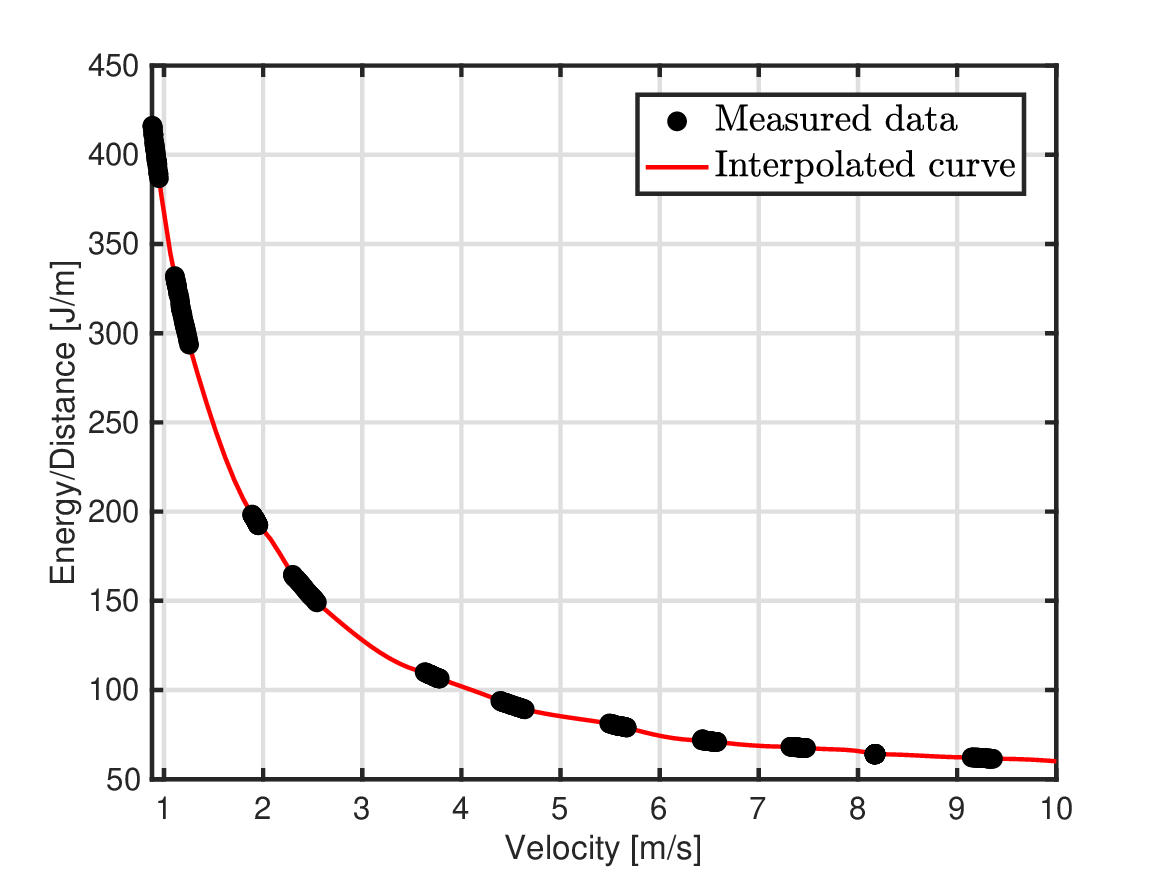}\\
  {\centering (b)}
\end{minipage}
\caption{Power/energy consumption model as a function of the constant velocity. (a) Power consumption for flight simulation. Dash lines are the interpolation for each dataset. (b) Energy consumption per unit distance. The interpolated curve is used for $P_{\rm{const}}$.}
\label{fig:energy_model}
\end{figure}

The cost $C_{\rm{E}}$ is quantified based on the power consumption model, which is developed by the data-driven approach \cite{Baek2019, Hong2020}.
The Gazebo simulator was employed again to measure the power consumption of the UAV during its flight.
The UAV flew horizontally with several command velocities of [1.0, 2.0, 4.0, 6.0, 8.0, 10.0] m/s.
The power consumption was calculated using the thrust forces of four motors of the UAV \cite{motor_spec}.
Each thrust was measured using the Gazebo force sensor plugin \cite{f_sens_plug}.
Figure \ref{fig:energy_model} (a) shows the measurement results for each velocity.
The plots for the constant velocity were recorded for at least 5 s, and the plots for acceleration/deceleration were measured during changing the command velocity.

Once the dataset is collected, the power consumption model for acceleration/deceleration and the energy consumption model per unit distance are learned based on the velocity of the UAV and given by the interpolated curves in Figure \ref{fig:energy_model} (a) and (b).
Then, $C_{\rm{E}}$ for a trajectory segment $(\bm{x}_{i}, \bm{x}_{i+1}, ... \bm{x}_{\rm{tip}})$ with a current velocity $V_{\rm{cur}}$ is calculated using the following equation:
\vspace{-2mm}

\footnotesize
\begin{eqnarray}
  \label{eq:energy_cons}
  C_{\rm{E}} = \frac{1}{P_{\rm{max}}}\left[\int_{t_i: V_{\rm{cur}}}^{t_k: V_{\rm{tmp}}} P_{\rm{acc/dec}} dt + \frac{P_{\rm{const}}}{V_{\rm{tmp}}}d\right],
\end{eqnarray}
\vspace{-2mm}

\noindent
\normalsize
where $P_{\rm{max}}$ is the maximum value of the power consumption for the normalization,
$P_{\rm{acc/dec}}$ denotes the power consumption model for acceleration and deceleration, which is given by the interpolated curves in Fig. \ref{fig:energy_model} (a),
$P_{\rm{const}}$ shows the model for the constant velocity,
$d$ is the distance of the trajectory segment.
The first term calculates the energy consumed from the time of $V_{\rm{cur}}$ to that of $V_{\rm{tmp}}$ with the maximum acceleration/deceleration $a_{\rm{max}}$.
$P_{\rm{const}}$ divided by $V_{\rm{tmp}}$ in the second term is expressed by the learned model in Fig. \ref{fig:energy_model} (b).
\vspace{-2mm}
\section{EXPERIMENTAL RESULTS}
\begin{table}[t]
\centering
\vspace{3mm}
\caption{Parameters used in trajectory planning simulation.}
\label{table:param}
\scalebox{0.8}{
\begin{tabular}{lccccccccc}
\hline
Parameter&$\alpha_{\rm{E}}$&$r_{\rm{max}} $& $r_{\rm{min}} $&$n_r$& $n_{\theta}$ &$V_{\rm{max}} $&$V_{\rm{min}} $&$P_{\rm{max}} $&$a_{\rm{max}} $\\
Value&1.0&41.0&5.0&6&12&10.0&1.0&600.0&1.0\\
Unit&-&m&m&-&-&m/s&m/s&W&s$^2$\\
\hline
\end{tabular}}
\end{table}

\vspace{-1mm}
We conducted two simulation experiments: a MATLAB-based simulation to verify the function of the proposed algorithm and a Software-in-the-Loop (SIL) simulation to validate it.
Table \ref{table:param} summarizes the parameters used for the simulations.
\vspace{-1mm}
\subsection{MATLAB-based Simulation}
\begin{figure}[!t]
\hspace{-3mm}
\begin{minipage}[t]{0.499\linewidth}
\centering
\includegraphics[width=1.01\linewidth]{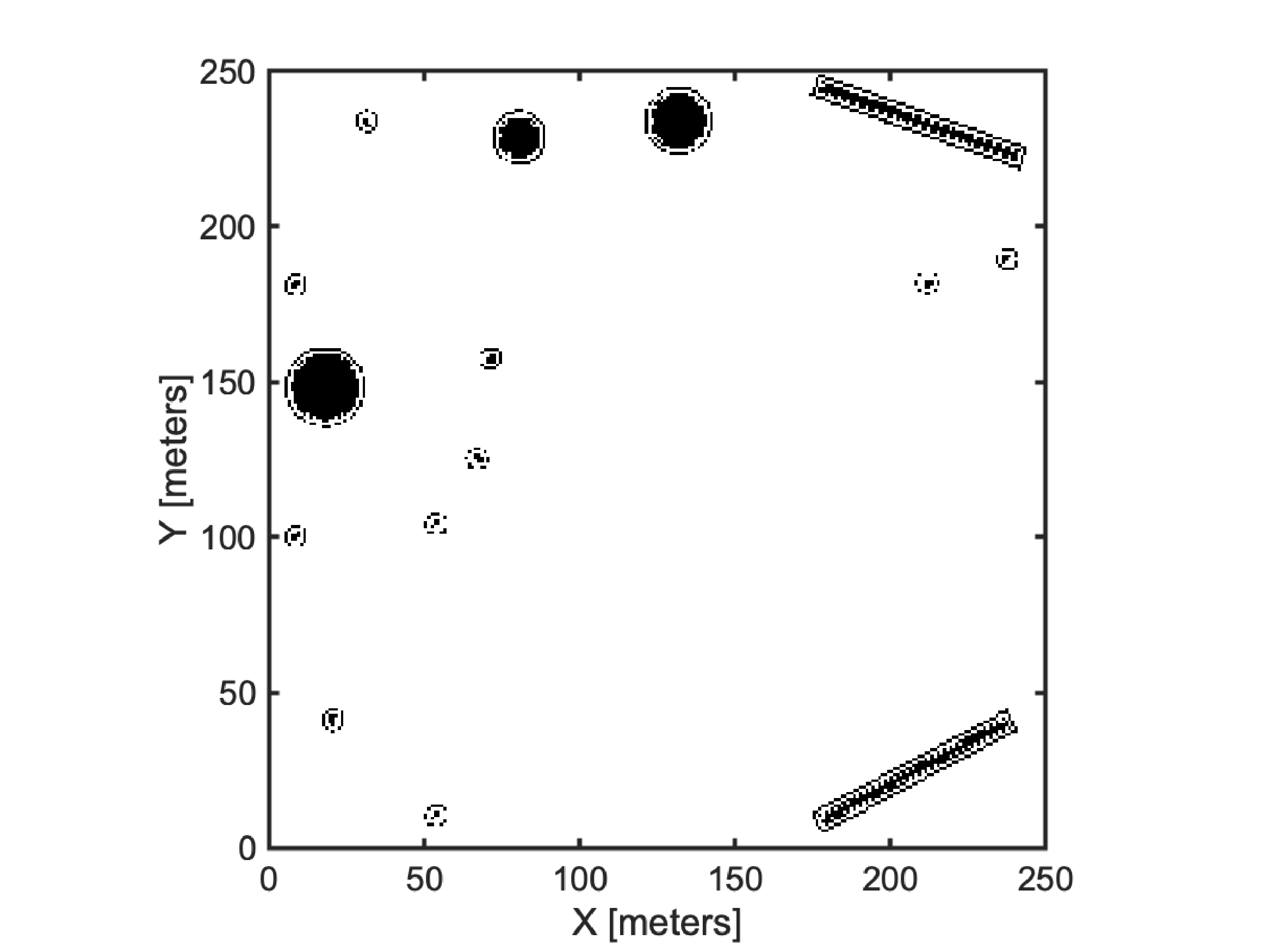}\\
  {\footnotesize(a) Raw map data for Scenario 1}
\end{minipage}
\begin{minipage}[t]{0.499\linewidth}
\centering
\includegraphics[width=1.01\linewidth]{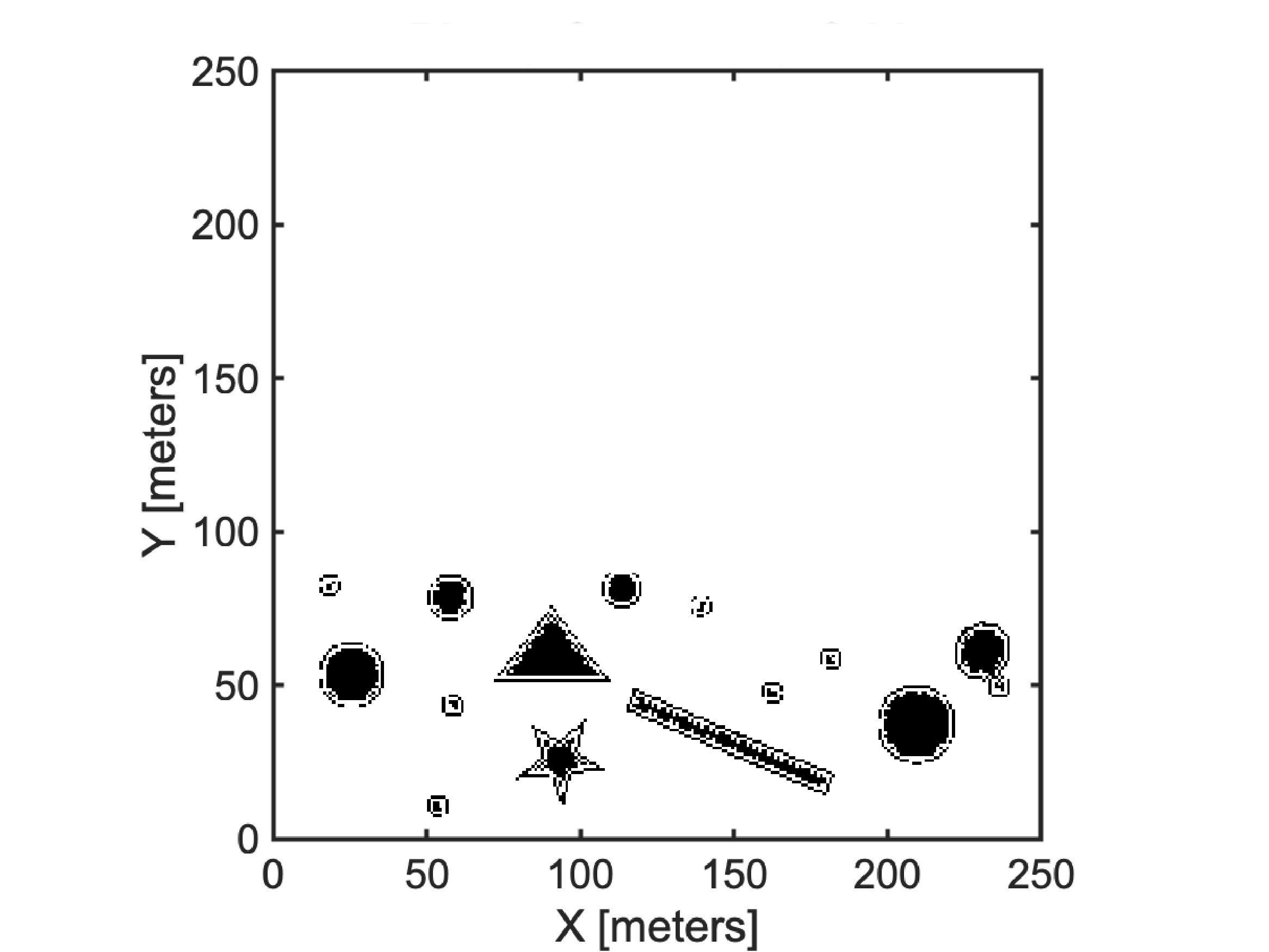}\\
  {\footnotesize(b) Raw map data for Scenario 2}
\end{minipage}
 \\
 \hspace{-3mm}
\begin{minipage}[t]{0.495\linewidth}
\centering
\includegraphics[width=1.0\linewidth]{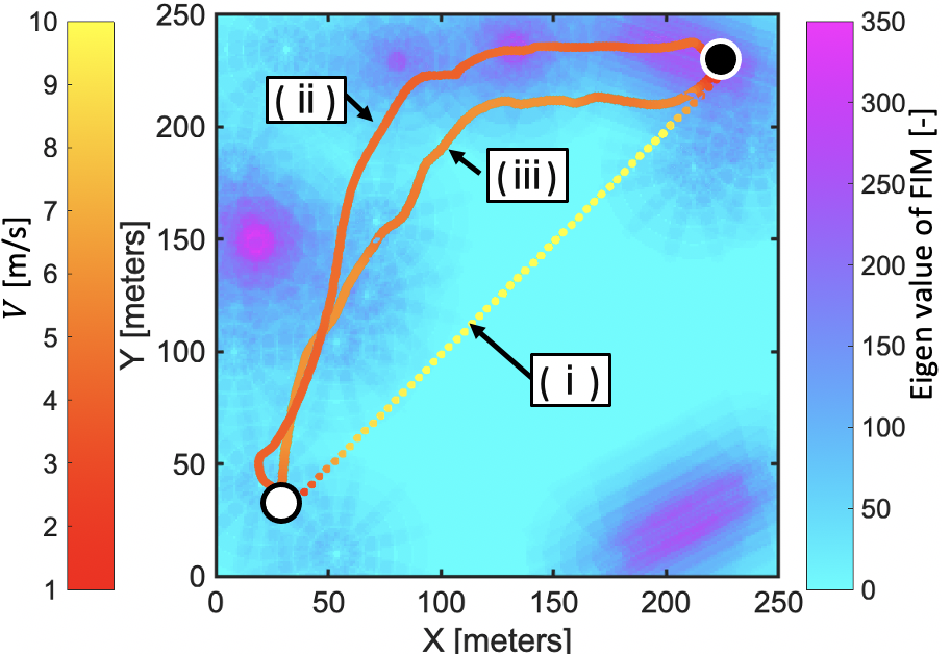}\\
  {\footnotesize(c) Planning result in Scenario 1}
\end{minipage}
\begin{minipage}[t]{0.495\linewidth}
\centering
\includegraphics[width=1.0\linewidth]{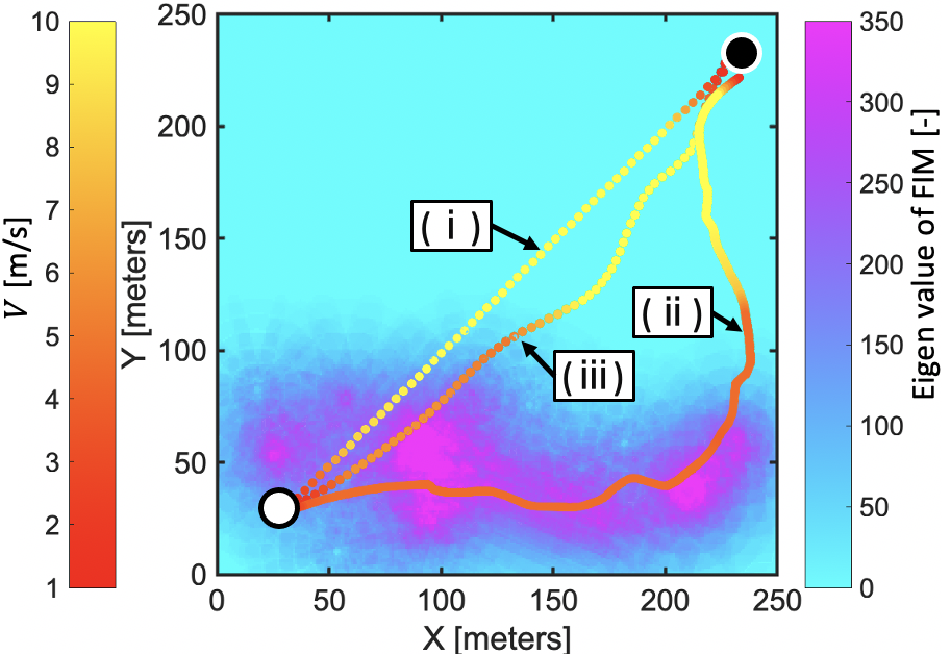}\\
  {\footnotesize(d) Planning result in Scenario 2}
\end{minipage}
  \vspace{-1mm}
\caption{Results of trajectory planning simulation. Black dots in the raw map are feature point for the SLAM. In the planning result, we illustrate the start and goal areas as white and black circles, respectively. (i), (ii), and (iii) depict the direct, high perception-quality, and proposed trajectory, respectively. The eigenvalue of the FIM is calculated at the velocity of 5 m/s for each position and represented by cyan and magenta gradient overlay: cyan (low perception quality) to magenta (high perception quality). Other color bar shows the velocity range of a UAV.}
\label{fig:traj_plan_result}
\end{figure}

\vspace{-1mm}
We highlighted the effect of the proposed planner with $\alpha_{\rm{P}} = 4$ through a comparison with a direct trajectory between the start and goal with $\alpha_{\rm{P}} = 0$ and a high perception-quality (HPQ) trajectory with $\alpha_{\rm{P}} = 1000$.
We intended the direct trajectory to be the most energy efficient and the HPQ trajectory to be the most sensitive to LiDAR measurements.
The planner was tested on two typical scenarios as shown in Fig.~\ref{fig:traj_plan_result} (a) and (b) and twenty trials because of the randomness of the RRT$^*$ algorithm.

The simulation results are summarized in Table \ref{table:eval1}.
The perception quality in the tables is calculated by the summation of the eigenvalue of the FIM throughout a trajectory.
The typical trajectory is also shown in Fig. \ref{fig:traj_plan_result} (c) and (d).
According to the comparison between the raw map and planning result, the area around feature points are assigned as higher FIM values, implying that the FIM calculation Eq. (\ref{eq:fi}) is applicable to the perception quality under heteroscedastic uncertainty.
Further, the proposed and HPQ trajectories encouraged the UAV to approach the area with higher FIM values, resulting in a higher perception quality than that of the direct trajectory.
Especially, we observed that the velocity tended to decrease where the perception quality was relatively large, while the UAV first flew where the FIM was small.
According to Fig. \ref{fig:lidar_exp_unc_model} (b), we deduce this is because the LiDAR measurement uncertainty for the FIM reduces as the UAV flies slowly.
Regarding total energy consumption, although the proposed trajectory required 6.9 and 4.2 kJ more than the direct trajectory for each scenario, it saved 9.1 and 7.3 kJ more than the HPQ trajectory.
Overall, this simulation result implies that the proposed planner enables the UAV to fly along a perception-aware trajectory under heteroscedastic uncertainty while maintaining energy efficiency.

\vspace{-1mm}
\subsection{SIL Simulation using Photorealistic Environment}
\begin{figure}[!t]
\centering
\includegraphics[width=0.7\linewidth]{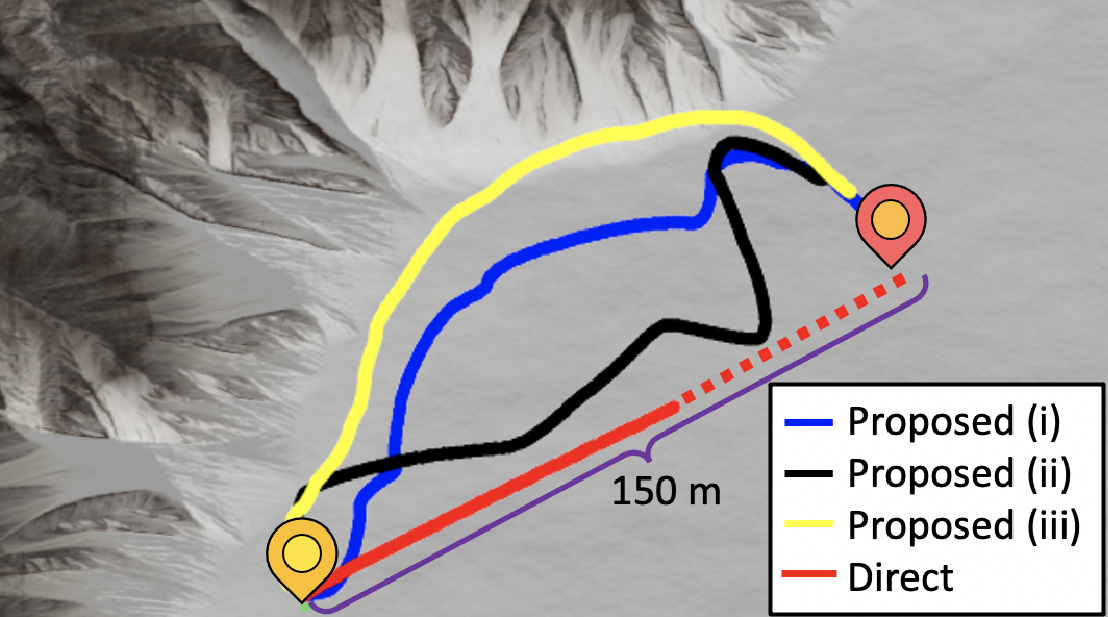}\\
  \caption{Results of Gazebo simulation for the validation of the proposed planner. Flown trajectories overlaid on the top view of the environment. The results of the proposed planner (blue) and a direct trajectory (red) are depicted. Red dash line represents the direct trajectory followed by the UAV after its positional error exceeds 30 m. We illustrate the start and goal areas as yellow and red pinpoints, respectively.}
\label{fig:valid_plan_result}
\end{figure}

We implemented the proposed planner on a robot operating system to validate it in the SIL simulation.
The proposed planner is iteratively executed using the latest scanned point cloud in a receding horizon manner until the UAV reaches the specified goal area.
We highlighted the effect of the proposed framework through a comparison with a direct trajectory with the maximum velocity ($\alpha_{\rm{P}} = 0$) using photorealistic 3D models referred from \cite{3denv}, as in \cite{Bartolomei2020}.
Additionally, we simulated three times to eliminate the randomness of the planner.
Figures \ref{fig:valid_plan_result} shows the results of the simulation.

Notably, the UAV with the direct trajectory attempted to move to the destination with $9.2$ kJ energy consumption.
However, its pose estimation error exceeded $30$ m in the middle of its flight, indicating that its flight failed.
Conversely, while our proposed planner needed $22.3 \pm 1.7$ kJ for its flight, it encouraged the UAV to approach feature-rich areas and succeeded in all flights, maintaining smaller final position error ($4.8 \pm 2.8$ m), which was only 3 \% of the distance between the start and goal area.
Overall, the simulation result validated that our proposed motion planner can work as real time motion planner and avoid the loss of UAV's localization.

\begin{table}[t]
\centering
\caption{Simulation results for Scenario 1 and 2}
\vspace{-3mm}
\label{table:eval1}
\scalebox{0.85}{
\begin{tabular}{|l|c|c|c|c|}
\hline
   &\multicolumn{2}{c|}{Energy} &\multicolumn{2}{c|}{Perception}\\
   &\multicolumn{2}{c|}{consumption [kJ]}& \multicolumn{2}{c|}{quality [-]}\\
  \hline
  Scenario & 1 & 2 & 1 & 2\\
\bhline{1.25pt}
    (i) Direct &20.5 & 20.5& 0.9$\times10^3$ & 0.1$\times10^3$\\
    (ii) HPQ &36.5$\pm$1.5&32.0$\pm$4.6&10.5$\pm$1.1$\times10^3$&9.4$\pm$3.0$\times10^3$\\
    (iii) Proposed &27.4$\pm$0.6&24.7$\pm$0.6&2.6$\pm$0.3$\times10^3$&1.7$\pm$1.6$\times10^3$\\
\hline
\end{tabular}}
\end{table}

\vspace{-2mm}
\section{CONCLUSION}
\vspace{-2mm}
In this study, we proposed perception-and-energy-aware motion planning for UAVs in GNSS-denied environments.
The LiDAR measurement uncertainty was learned as a function of the horizontal velocity of the UAV.
The online planning exploited the model to estimate the perception quality.
Simulation experiments verified the FIM can be applicable to the perception quality under heteroscedastic uncertainty.
Further, the proposed planner enabled the UAV to approach the feature-rich area with appropriate velocity commands in terms of perception quality and energy efficiency, resulting in the avoidance of the loss of localization.

Future studies will possibly include the improvement of the accuracy of the high-fidelity simulator.
To implement the proposed planner on our UAV testbed, the accurate uncertainty model is necessary; however, in field experiments, it is almost impossible to measure LiDAR uncertainty as discussed.
Thus, modeling a higher fidelity simulator is a solid approach.

\bibliographystyle{IEEEtran}

\end{document}